%% file: bare_conf_PX.tex
\documentclass[conference]{IEEEtran}
\usepackage[pdftex]{graphicx}
\usepackage{amsmath}
\usepackage{algorithmic}
\usepackage{siunitx} 

\usepackage{multirow}
\usepackage{booktabs}
\usepackage{makecell}
\usepackage[dvipsnames]{xcolor}

\begin{document}
%
\title{Plasticity-Enhanced Domain-Wall MTJ Neural Networks for Energy-Efficient Online Learning }



%

\author{\IEEEauthorblockN{ Christopher H. Bennett$^1$, T. Patrick Xiao $^1$, Can Cui $^2$, Naimul Hassan$^3$, Otitoaleke G. Akinola$^2$, \\
Jean Anne C. Incorvia$^2$, Alvaro Velasquez$^4$, Joseph S. Friedman$^3$ , and Matthew J. Marinella$^1$ }
\IEEEauthorblockA{$^1$ Sandia National Laboratories, Albuquerque, NM 87123 \\
$^2$Department of Electrical and Computer Engineering, University of Texas at Austin, Austin, TX 78712\\
$^3$Department of Electrical and Computer Engineering, University of Texas at Dallas, Richardson, TX 75080\\
$^4$ Information Directorate, Air Force Research Laboratory, Rome, NY 13441\\
\{cbennet, mmarine\}@sandia.gov, joseph.friedman@utdallas.edu, alvaro.velasquez.1@us.af.mil}}


\maketitle

\begin{abstract}
Machine learning implements backpropagation via abundant training samples. We demonstrate a multi-stage learning system realized by a promising non-volatile memory device, the domain-wall magnetic tunnel junction (DW-MTJ). The system consists of unsupervised (clustering) as well as supervised sub-systems, and generalizes quickly (with few samples). We demonstrate interactions between physical properties of this device and optimal implementation of neuroscience-inspired plasticity learning rules, and highlight performance on a suite of tasks. Our energy analysis confirms the value of the approach, as the learning budget stays below $20 \mu J$ even for large tasks used typically in machine learning.
\end{abstract}


%
\IEEEpeerreviewmaketitle

\section{Introduction}
The drive towards autonomous learning systems requires computing tasks locally or \textit{in-situ}, defraying rising energy costs due to inefficiencies in the modern computer architecture \cite{horowitz20141}. A variety of emerging non-volatile memory devices, such as phase-change materials, filamentary resistive RAM, and magnetic memories (spin-transfer-torque-RAM (STT-RAM) and spin-orbit-torque-RAM (SOT-RAM)), may implement this vision. Critically, emerging devices can perform not only data storage but complex physics-powered operations such as vector-matrix multiplies (VMMs) when densely wired \cite{burr2017neuromorphic}.

The workhorse algorithm in AI workloads is backpropagation of error (BP). BP relies upon a teacher signal supplied to all layers and the storage of high-quality gradients on each layer during the parameter update phase \cite{rumelhart1995backpropagation}. In contrast, competitive learning or adaptive resonance methods provide labels sparsely, \textit{e.g.} only to some parts of the system; the rest learn according to internally adaptive units and/or dynamics \cite{grossberg1987competitive}. Competitive learning relies upon the winner-take-all (WTA) motif, a cascadable non-linear operation that can be used to build deep systems, just as perceptrons can be used to build multi-layer perceptrons (MLP) \cite{maass2000neural,maass2000computational}. Original proposals for building WTA circuits relied upon a chain of inhibition transistors \cite{mead1991winner}. Analog and digital WTA or spike feedback CMOS systems have been realized \cite{ramakrishnan2013vector,park20197}, and conceptual proposals for WTA systems using emerging devices exist \cite{truong2016ta,wu2014analysis}. However, these works either do not discuss scalable (local) learning rules that might lead to large-scale WTA systems, or do not adequately benchmark against state-of-the-art tasks in the machine learning field . In order to implement efficient WTA learning, we draw upon the spike-timing-dependent plasticity (STDP) rule, a primitive predictive/correlative engine \cite{rao2001spike}. As in \cite{vincent2015spin}, we implement STDP and WTA learning together with emerging memory, however our chosen synapses are analog and, as in \cite{krotov2019unsupervised}, we closely study neuronal behavior/interactions to implement optimal competitive learning with hidden units.

\begin{figure}[!t]
\centering
\includegraphics[width=0.6\columnwidth]{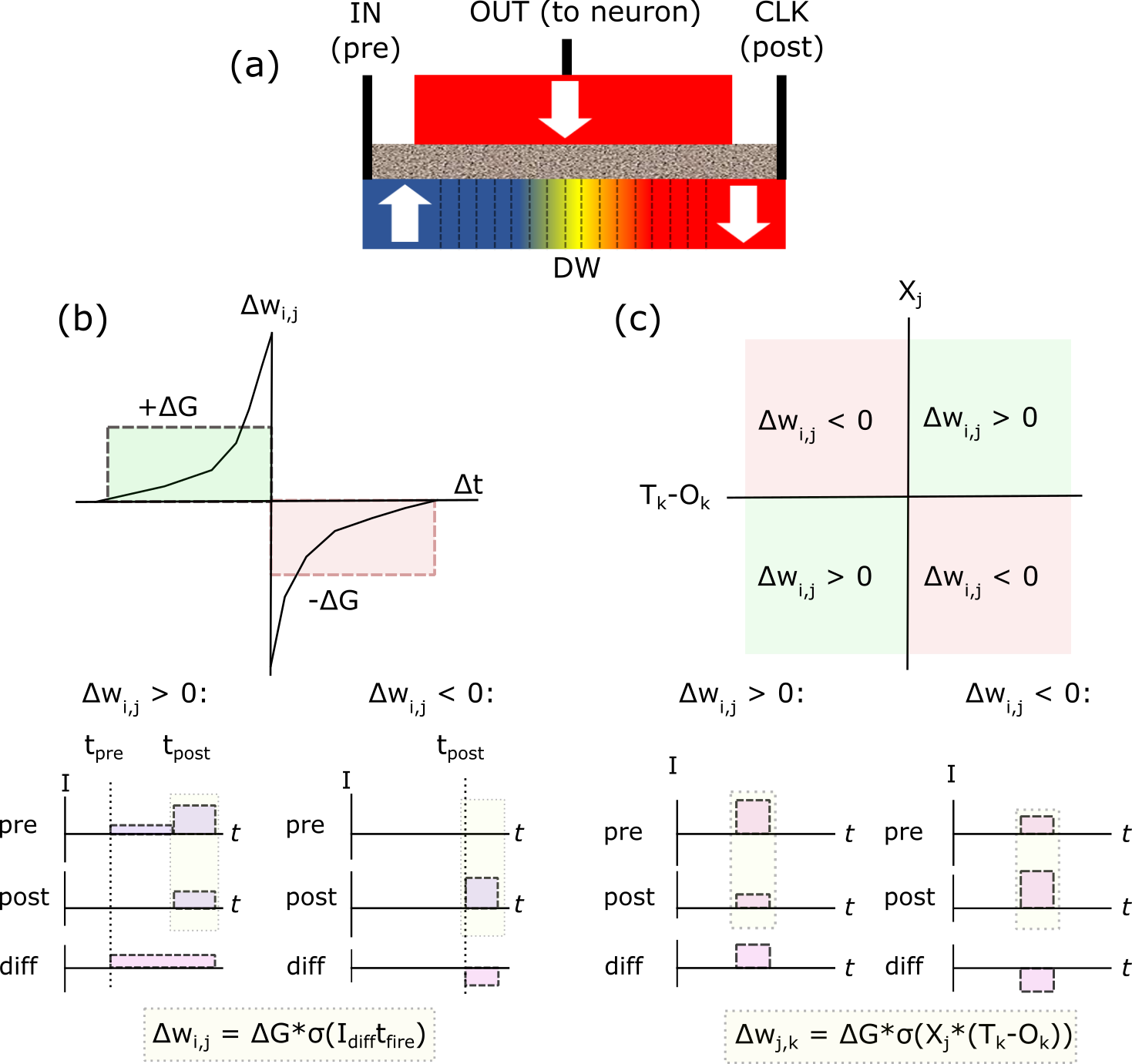}
\caption{(a) An illustration of the DW-MTJ analog synapse. (b) abstract learning rule (top) and its temporal implementation using physical currents in the system for the WTA competitive learning system (first layer) and (c) the same for the supervised learning system (second layer).}
\label{fig:rule}
\end{figure}

Our chosen analog memory is the three-terminal magnetic-tunnel-junction (3T-MTJ) device. These devices: 1) achieve high switching efficiency due to the SOT interaction at input/output terminals; 2) possess a non-volatile state variable, a domain-wall interface (DWI) moving through a soft ferromagnetic track; 3) can be dually utilized as a synapse, holding an internal conductance state $G_\text{MTJ}$ when the output terminal is long, or implement the neuron function, when the track is long. In the former case, domain wall synapses notably possess good energy footprint and advantageous operation on neural network tasks in comparison to other nanodevice synaptic options \cite{kaushik2019comparing}. In the latter case, 
assuming tight spacing lateral inhibition exists between neighboring DW-MTJ neuron tracks, and the physics-derived leak function can be used to implement rapid inference operations given pre-trained weights \cite{hassan2018magnetic}. In this work , we describe an efficient combination of unsupervised (WTA+STDP) and supervised (label-driven) learning in an all-DW-MTJ device array that approaches BP-level performance and remarkable energy efficiency on difficult tasks.

\section{Operation of Nanomagnetic WTA Primitive}
Our system relies upon three operations 1) Inference: a vector-matrix-multiplies on clustered weights $\mathcal{W}_c$ generate post-synaptic outputs. 2) Domain-Wall Competition : A dynamic step whereby interacting neuron units evolve according to post-synaptic inputs (a vector of currents $\vec{I_\text{post}}$), as well as the behavior or nearby neighbor units, according to a physics-informed model. 3) Learning/Programming: An update step where weights $\mathcal{W}_c$ are updated according to a simplified version of the spike-timing-dependent plasticity (STDP) rule; neurons implement different hidden statistical models of the input \cite{kappel2014stdp}. These stages are progressively implemented in the unsupervised phase (label-free). Once $N_{us}$ unlabeled examples from the training set have been seen, weights are frozen and a least-mean-squares (LMS) filter is progressively built in a second weights matrix $\mathcal{W}_s$ using $N_s$ labeled data points.

\begin{figure}[!t]
\centering
\includegraphics[width=0.6\columnwidth]{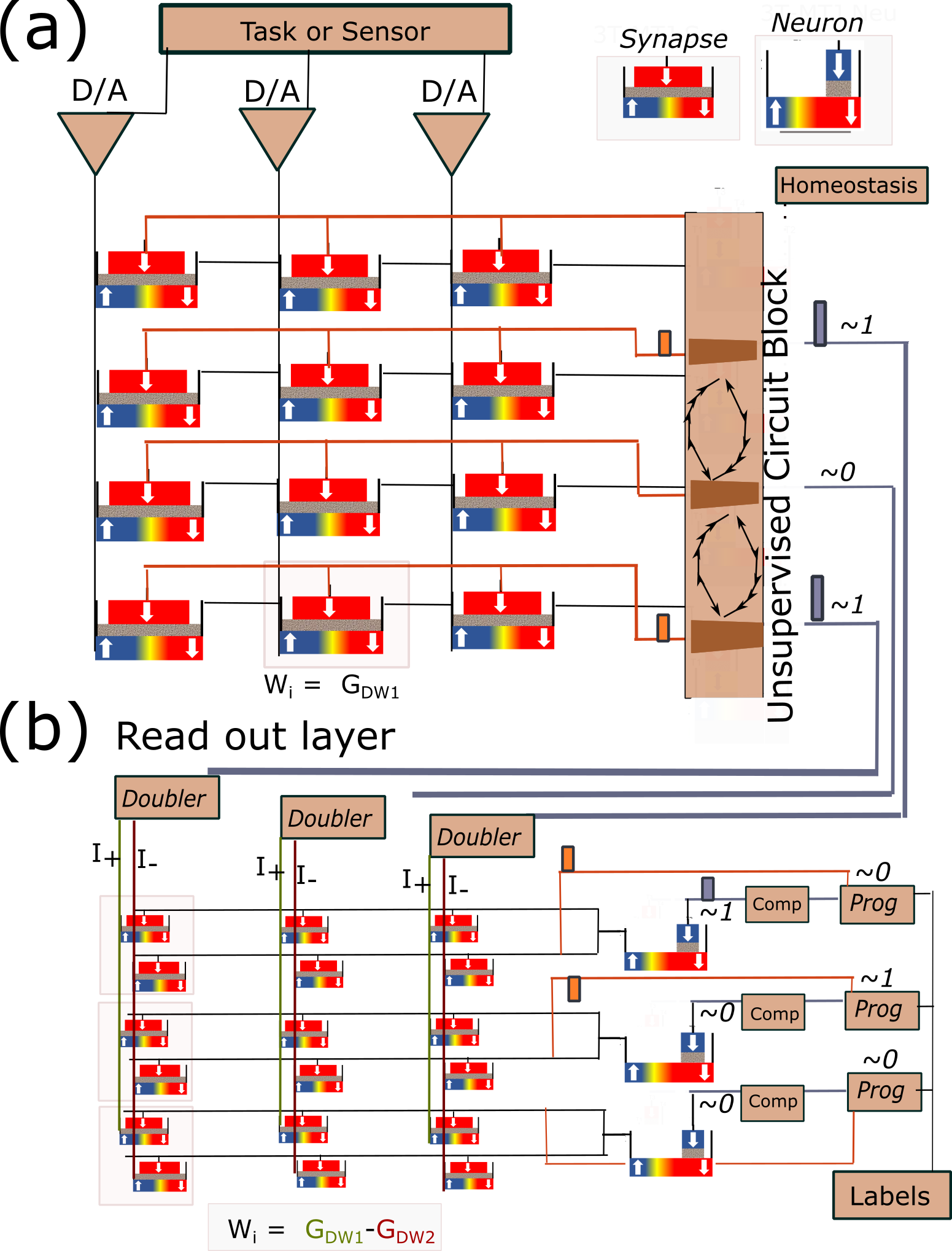}
\caption{(a) The clustering layer(s) takes as input and output analog currents. Following spiking of DW-MTJ neurons, back-propagated spikes return to the row implementing clustering. The clock terminal is typically grounded but may be connected to a homeostatic regulation signal. (b) The read-out layer takes spikes from the hidden layer doubled using current mirrors and encodes real-valued weights with 2 MTJ synapses per cell. Programming follows comparison of spikes to expected values.}
\label{fig:system}
\end{figure}

\subsection{Details of Lateral Inhibition Model}
As in \cite{hassan2018magnetic}, the dependence of a magnetic stray field's transverse (vertical) component impinges upon that of neighboring wires. This can be described by: 
\begin{equation}
H_z = \frac{4M_s}{\pi}(arctan\frac{2s+3w}{t} - arctan\frac{2s+w}{t})
\end{equation}
based on \cite{engel2005calculation}. Here, $M_s$ is the magnetic saturation field set at 1.6T, $w$, $t$, and $s$ are width, thickness of the track and inter-wire spacing respectively. When $H_z$ is in the proper range, it can effectively reduce DW velocity $v$. Instead of rigorously calculating $H_z$ in the neural simulator, we focus on an ensemble parameter $ \gamma $ that modifies naive, current-dominated DW motion $v_0$: 
\begin{equation}
\gamma = \frac{v_{0}(w,t,I_{in}) -v_\text{inhib}(w,t,I_{in},s)}{ v_0(w,t,I_{in})} 
\end{equation}
This ratio captures the predominance of current-driven vs. coupled (field-driven) DW behavior. At very low $\gamma$, field influences are negligible; at $\gamma \approx 0.5 $, coupling is intermediate, and current and field DW influences are mixed; as $\gamma$ approaches $1 $, neighbor field effects outweigh the influence of input current. Physically, the spacing $s$ can vary between 10nm and 150nm spacing in order to reflect a full spectrum of coupling strength. However, $\gamma$ may not evolve linearly in this regime, as demonstrated in \cite{cui2019maximized}.

\subsection{Details of Analog Plasticity Model}
As in \cite{currivan2012low}, the number of weights given a domain wall length $L_\text{dw}$, track width $w$, and length of output MTJ terminal $L_\text{mtj}$ (where the analog conductances are realized) is
\begin{equation} \label{eq:N_w}
n_w \leq \frac{L_\text{mtj}}{L_\text{dw}} \approx 4\frac{L_\text{mtj}}{w}
\end{equation}
Given $w=32 nm$, 6 bits could be implemented given an output port length of $512 nm$. Analog weights can be implemented with the use of notches for precise control and non-linearity \cite{akinola2019three}, or can be obtained intrinsically via fine current controlled pulses. Due to DWI momentum effects, notch-free systems will typically require greater output/synapse length.

During plasticity events, differences in currents between synaptic input and output 3T-MTJ ports determines the motion of the DWI modulating $G_\text{MTJ}$. As in Fig. \ref{fig:rule}, the circuit potentiates the synapse/increases the conductance when the two currents are coincident and depotentiates the synapse/decreases the conductance when they are not. This implements an approximate version of Hebbian/anti-Hebbian learning , or approximate STDP (hereafter $A-STDP$). The teacher signal implementation relies upon DW-MTJ neurons being connected backward to the synaptic devices of that layer , as in the orange wires shown in Fig. 2(a). Further electrical details on the scheme are given in \cite{alvarounsupervised}.

\subsection{Integration with Companion Supervised Learning System}
A WTA primitive can be difficult to interface, leading to the desire to efficiently combine unsupervised and supervised sub-systems \cite{querlioz2012bioinspired}. In our case, the results from the competitively learning DW-MTJ system are forward-propagated to a supervised learning layer that is constructed additionally from DW-MTJ synapses and neurons, as shown in Fig. \ref{fig:system} and first suggested in \cite{bennett2019semi}. This system contains $2MN$ total DW synapses to encode both positive and negative weights, where $M$ is the number of hidden nodes and $N$ is the label-applied terminal set of neurons. We have considered two possible strategies for the supervised learning policy. The first sign-based learning policy can be implemented with great energy efficiency in neuromorphic hardware \cite{thakur2015online}, and reduces to:
\begin{equation}
\Delta W_{j,k} = \Delta G \sigma ( X_j (T_k-O_k)) ,
\end{equation}
where $X_j$ is the input from hidden neuron $j$, $O_k$ is the output at the $k^\text{th}$ terminal neuron, $T_\text{k}$ is the target (correct) label, $\sigma $ is the sign function and $\Delta G$ is the unit of conductance change per update. The second policy, softmax learning, requires an analog computation but can achieve superior results in machine learning contexts. Given the original post-synaptic update $Y_{k}$ , the softmax function is computed subsequently.
Weights are ultimately updated according to $ \Delta w_{j,k} = -\eta \delta_{j,k}$, given a learning rate $\eta$, and $ \delta_{j,k} = H_j(O_k-T_k)$ following the cross-entropy formulation , where $H_j$ is the pre-synaptic activation values of that layer $j$, as in \cite{bennett2019contrasting}. 

\section{Description of Data Science Tasks}
We consider three tasks: 1) the Human Activity Recognition (HAR) set of phone sensor data (\textit{e.g.} body acceleration, angular speed). There are 5 classes of activity (standing, walking, etc), 21,000 training and 2,500 test examples of dimension $L=60$ \cite{anguita2013public}. 2) the MNIST database of hand-written digits, which includes 60,000 training and a separate 10,000 test examples, at $L=784$ \cite{lecun2010mnist}. 3) The fashion-MNIST (f-MNIST) database, which is of same dimensionality as 2), represents items of clothing (sneakers, t-shirt, etc) and is notably less linearly separable than either of the previous tasks \cite{xiao2017fashion}. 

\section{Performance on Tasks}
\subsection{Parameters for successful clustering}
\begin{figure}[!t]
\centering
\includegraphics[width=0.625\columnwidth]{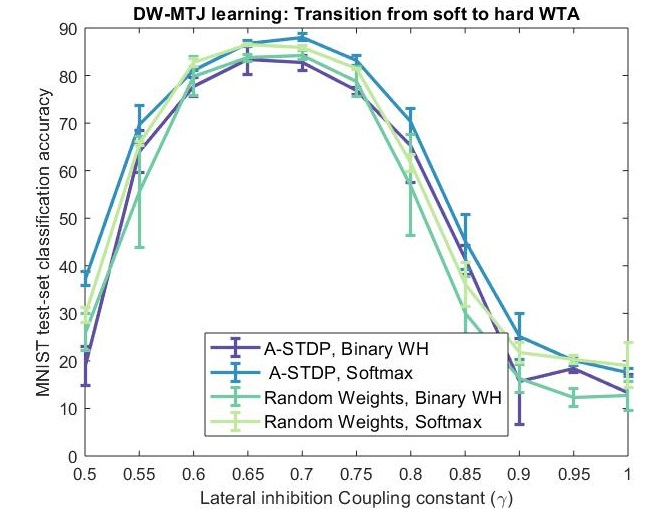}
\caption{Calibration of  $\gamma$ ; MNIST, $M = 600$ ,  $N_{us} = 1000$, $N_s=30000$.}
\label{fig:cc}
\end{figure}

For correct clustering system operation, the most critical parameter tends to be the coupling parameter $\gamma$. As visible in Fig. \ref{fig:cc}, while the intermediate/low amount of stray field interaction (over-firing) and dominant stray field interaction (under-firing) both do poorly, the high-intermediate level of interaction in which current matters but is outweighed by locally dominant neighbors results generalizes properly. Computationally, this suggests an intermediate point between 'hard' WTA (in which one or close to one neurons fire) and 'soft' WTA (in which most neurons fire) best implements clustering and forces a useful hidden representations of the input dataset.

\begin{table}
\caption{Classification and Regression Task Performance}
\label{Table1}
\centering
\begin{tabular}{@{}l*{4}{c}@{}}
\toprule
\text{Task} & \multicolumn{4}{c@{}}{\text{Learning Style}}\\
\cmidrule(l){2-5}
& Random, Ana-BP & STDP, Bin-BP & STDP, Ana-BP \\
\midrule
\makecell[l]{HAR} & $96.13 \%$ & $95.83 \% $ & $97.93 \% $& \\
\makecell[l]{MNIST} &$93.52 \% $& $93.12 \% $ & $94.92 \% $ & \\
\makecell[l]{f-MNIST} & $76.52 \% $ & $77.52 \% $& $79.52 \% $ &\\
\bottomrule
\end{tabular}
\end{table}

\begin{figure}[!t]
\centering
\includegraphics[width=0.9\columnwidth]{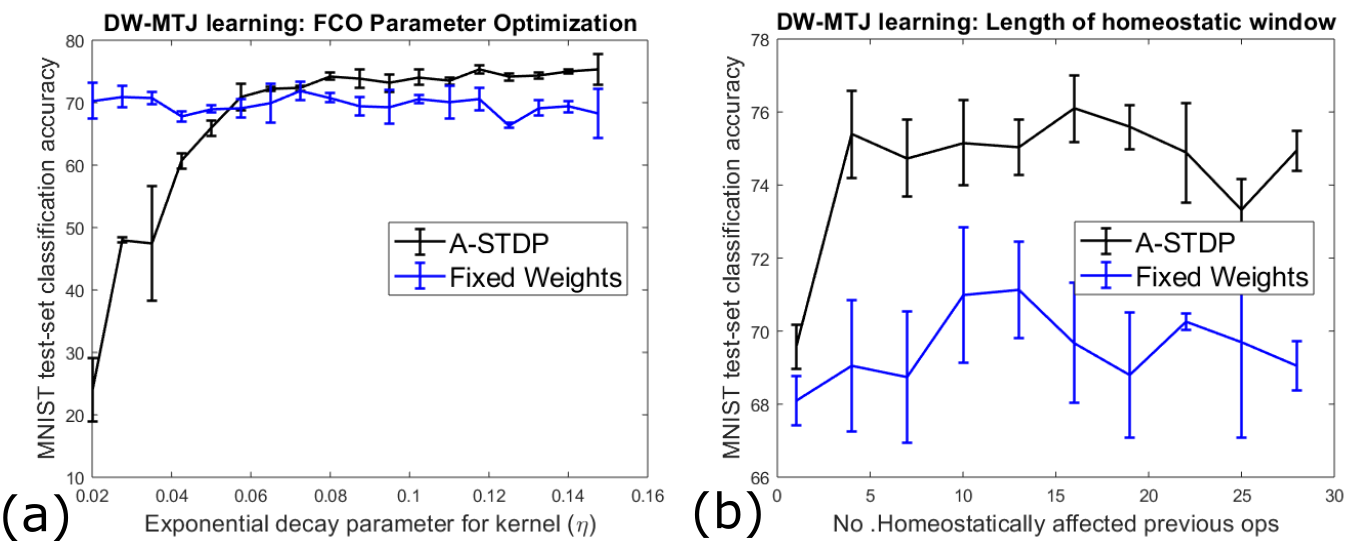}
\caption{Rank order filtering (a) and homeostatic delay mechanism (b) contribution to competitive learning with DW-MTJ neuron devices on the MNIST task. Simulated systems had $M=200 $ hidden layer neurons, given $N_{us} = 1000$ clustering samples, and $N_s=30000$ supervised samples.}
\label{fig:params}
\end{figure}

Next, we evaluate how critical two common enhancements to standard WTA operation -- homeostasis \cite{querlioz2015bioinspired} and rank-order coding \cite{bhattacharya2010biologically} -- are to strong performance in the hidden layer. Fig. \ref{fig:params} shows that these two operations are also important. In the case of homeostasis, we find that a small number of homeostatically inhibited time steps provides this benefit already, and a great deal of fine-tuning is not needed. A similar result is obtained for order coded learning, where a sufficiently large exponent is needed to clip the updates to a reasonable number of total neurons firing. Note that when this parameter is very low, the hidden layer tends to again over-fire and redundantly sample. Since correct values of $\gamma$ also naturally clip the total number that can fire, this suggests that the poor a-STDP results in Fig. \ref{fig:params}(a) are unlikely.

\begin{figure}[!t]
\centering
\includegraphics[width=0.9\columnwidth]{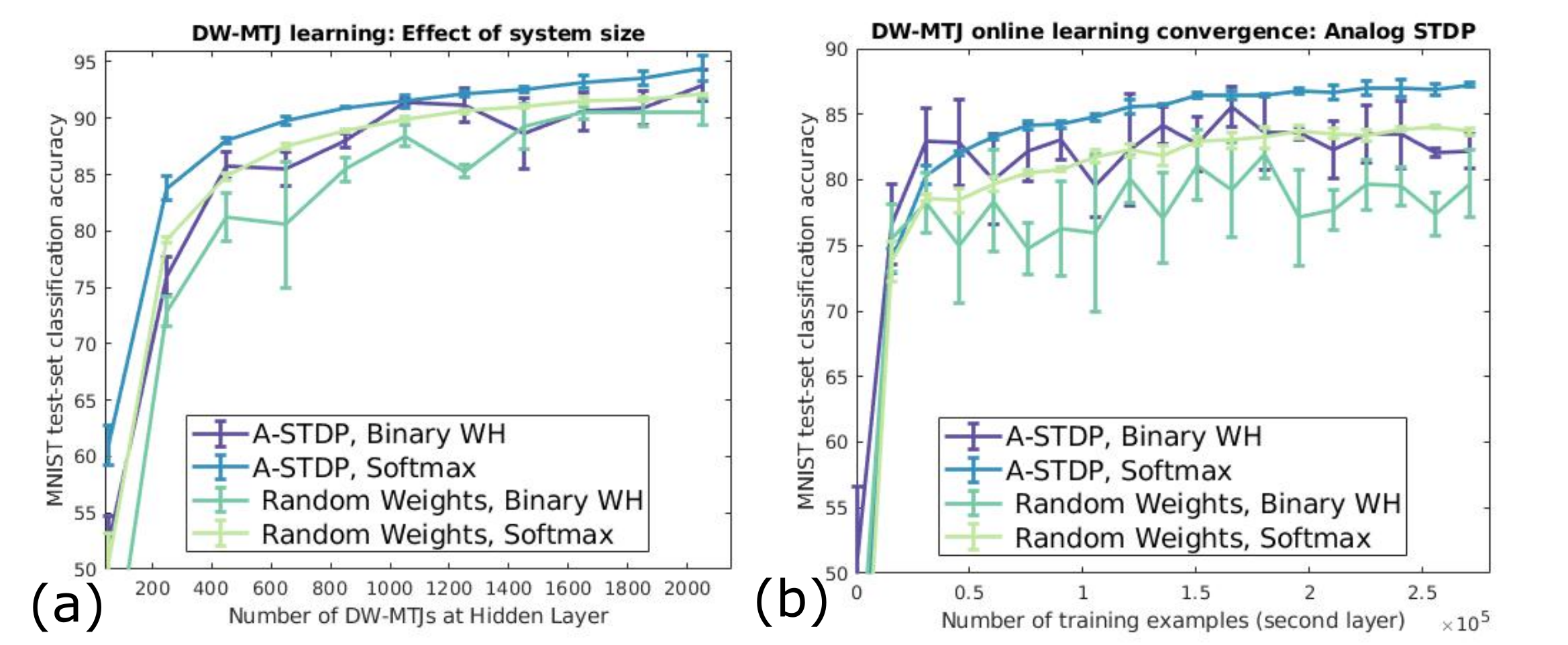}
\caption{The effect on MNIST classification performance of (a) the total number of competing hidden layer units and (b) the number of samples provided to the supervised layer to read out the results of the clustering operation. In (b), there are $M=400$ hidden-layer units.}
\label{fig:conv}
\end{figure}

\subsection{Dimensional and learning set requirements}
Fig. \ref{fig:conv} illustrates performance on MNIST task as a function of competing units $M$ and number of supervised training samples given a properly calibrated hidden layer. Ultimately, $~94.5\%$ classification on the test-set is achieved when using ana-BP in the second layer with only $N_s=30000$ examples drawn from the training set (but with a fairly large $M=1200$). Table 1 summarizes the top results for the other two tasks. For HAR, $~97\%$ is reached given $N_s=15000$ and $M=600$; f-MNIST requires $M=1800$ and $N_s=60000$.This suggests the current design is adequate on more separable tasks, while deeper networks may be required to prevent unacceptable system size blow-up on very non-separable (difficult) ones. These are notably low numbers for the total number of labeled data points presented; a modern memristive MLP requires many multiples of the task set, e.g. 200-500k samples for MNIST or f-MNIST \cite{kataeva2015efficient,bennett2019contrasting}, and achieves ~ 96\% on MNIST and 81 \% on f-MNIST. Thus, our present results are very slightly inferior to BP. However, as in Table 1, clustering outperforms the random weights system definitively, given the more robust learning procedure in the read-out layer.

\begin{figure}[!t]
\centering
\includegraphics[width=0.9\columnwidth]{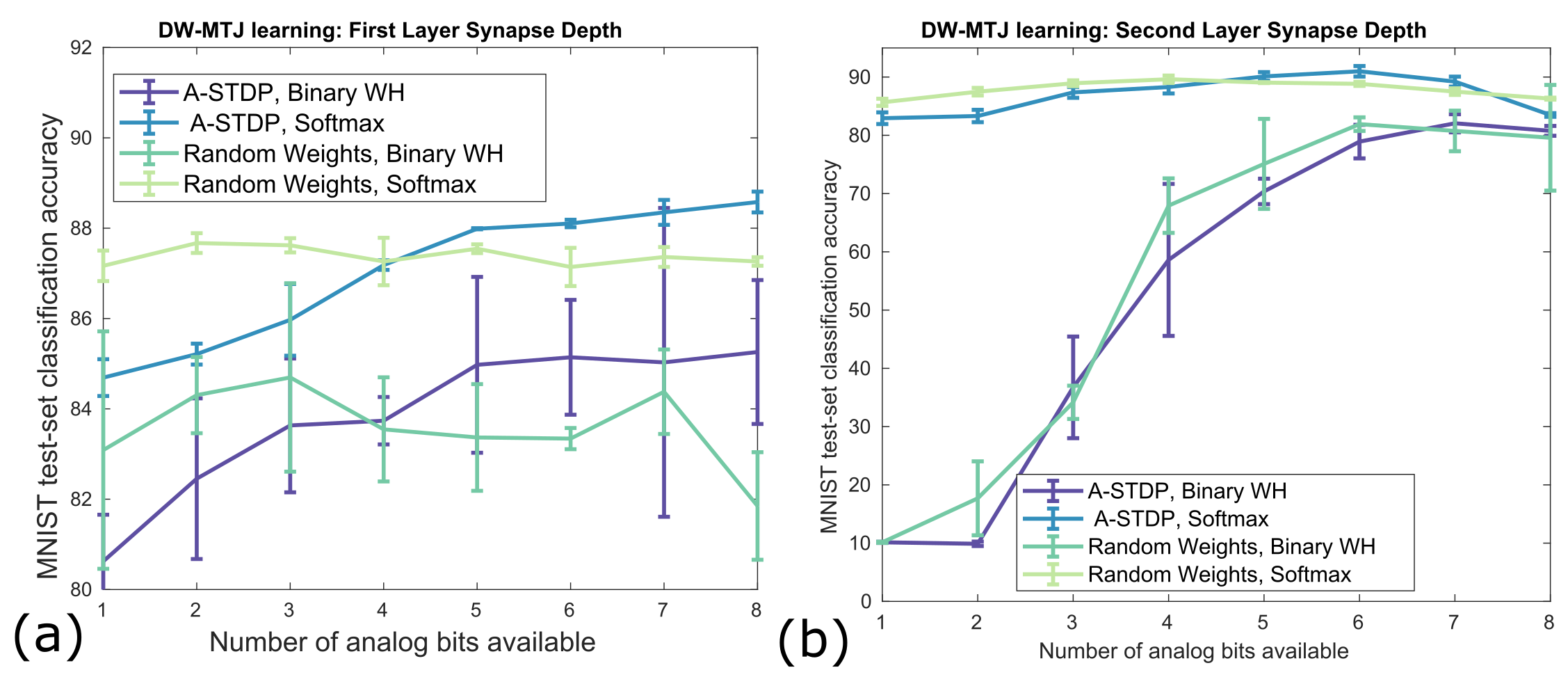}
\caption{Effect of writable space of $G_\text{MTJ}$ for the (a) first/clusteirng layer and (b) second/supervised layer; synapse depth is 6 bits in the other layer. For both, the task is MNIST and $M=500$, with 30,000 training samples.}
\label{fig:syn_depth}
\end{figure}

\subsection{Resilience to Intrinsic Physics Effects in System}
\begin{figure}[!t]
\centering
\includegraphics[width=0.9\columnwidth]{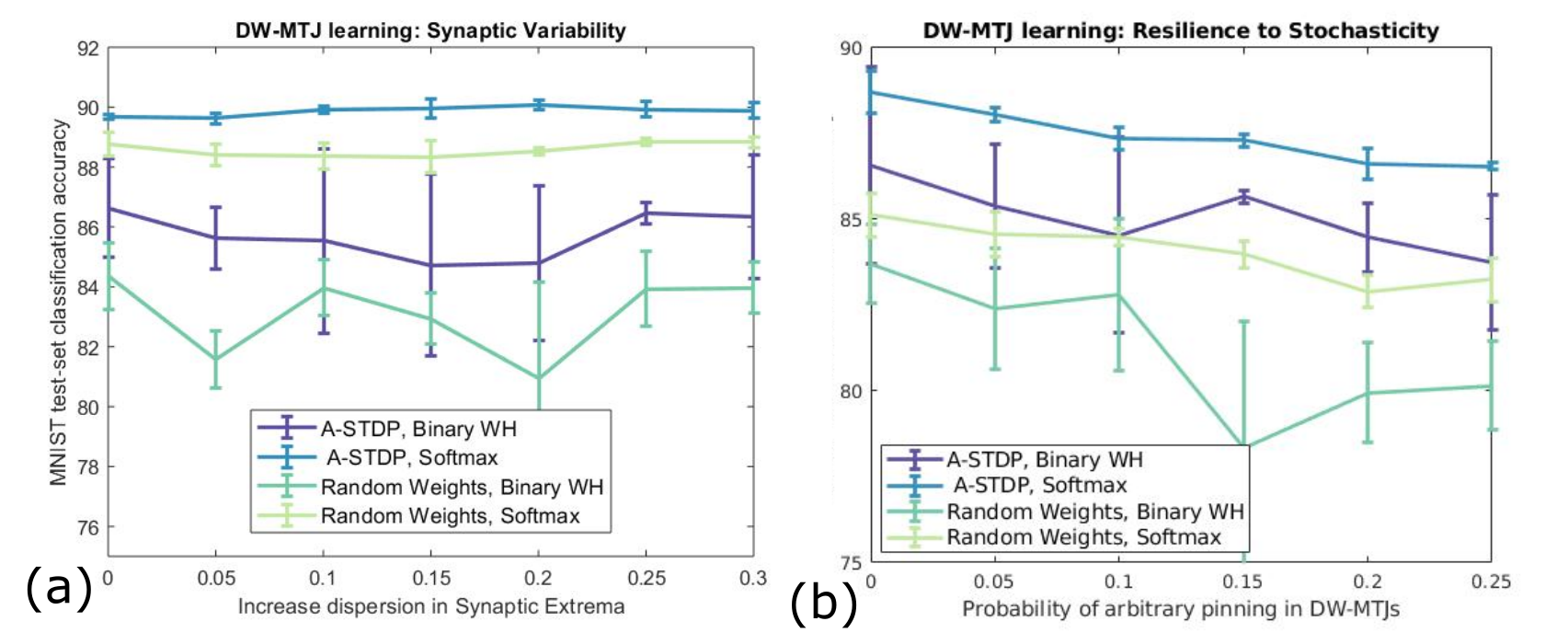}
\caption{(a) The effect of increasing variability of maximum and minimum states of DW synapses ($G_\text{on} $, $G_\text{off}$) in first layer\textit{e.g.} TMR variation. (b) The effect of random DW pinning. For both cases, $M=600$, 1000 clustering, 30000 training examples given on the MNIST task.}
\label{fig:var_stoch}
\end{figure}

Several issues may occur in the physical learning system which are non-ideal: a) synapse-level coarseness, \textit{e.g.} limited resolution of synapses; b) synapse-level process-induced variation at the output MTJ cell (which creates different $G_\text{On}/ G_\text{Off}$ states and TMR ratio); c) neuron-level stochastic effects due to natural fractal edge roughness in DW-MTJ nanotracks \cite{dutta2017spatial} which can cause a neuron, at a given clustering timestep, to fail to compete/fire. For coarseness, Fig. \ref{fig:syn_depth}(a) shows that $\mathcal{W}_c$ requires 4 bits per synapse to outperform random weights , regardless of second layer policy; performance continues to increase with more resolution, leveling off at 7-8 bits. Meanwhile, the supervised layer is sensitive to synaptic depth when using the binary BP rule but insensitive to it when using the analog rule- regardless of first-layer weight style. Next, Fig. \ref{fig:var_stoch}(a) shows that the clustering operation is almost unaffected by synapse-level variability. Finally, Fig. \ref{fig:var_stoch}(b) shows the effects of arbitrary domain wall pinning are significant and linear. If around $5 \%$ of neurons do not fire at any given clustering step, $1.0-2.5\% $ accuracy is lost. However, the effect of random pinning is negligible when not in ultra-low current operation.

\section{Energy Footprint of Proposed Systems}
\begin{figure}[!t]
\centering
\includegraphics[width=0.9\columnwidth]{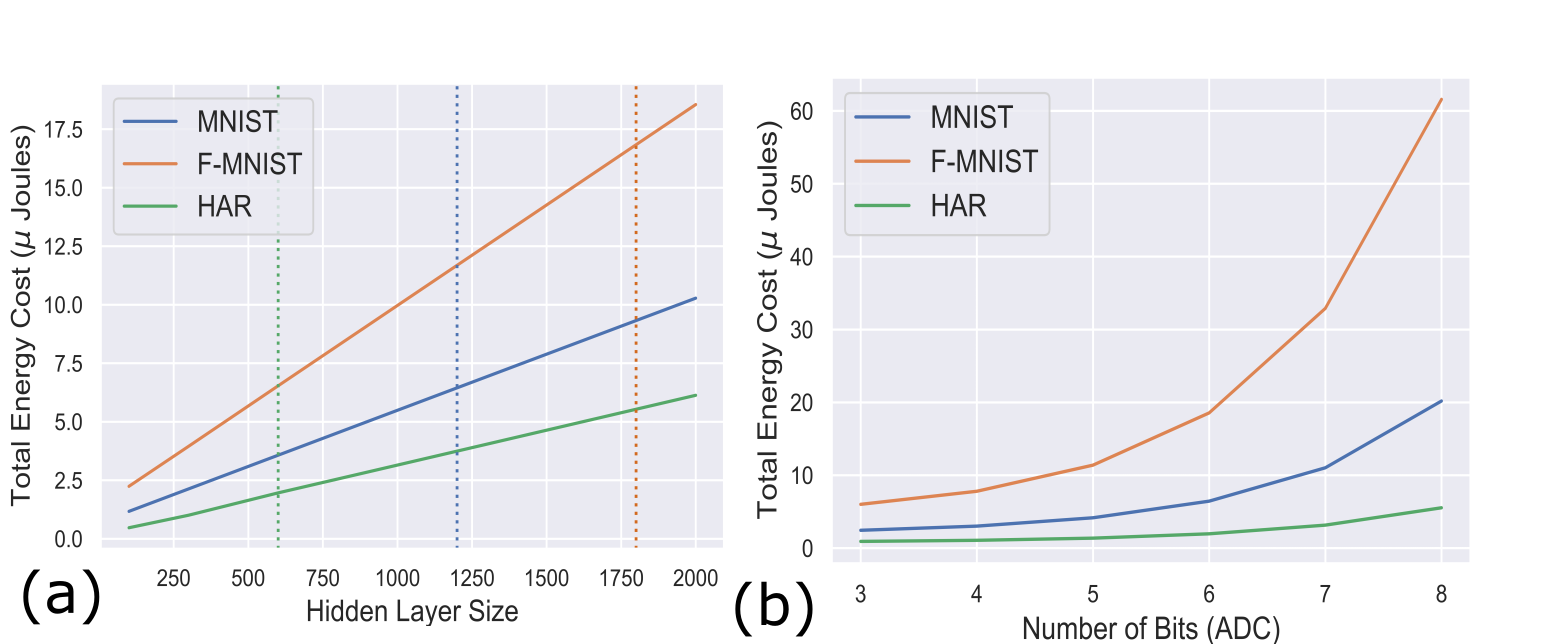}
\caption{Dependence of energy footprint given (a) hidden layer dimension $M$ assuming 6 bits in the ADC, and (b) the ADC bit resolution, given the $M$ values noted in (a) as the dotted (vertical) lines. 
}
\label{fig:energy}
\end{figure}

Drawing on methodology in \cite{bennett2019contrasting}, \cite{parmar2018design}, and \cite{marinella2018multiscale}, we estimate the energy overhead for the entire online learning procedure. On the device level, we have assumed that on average average $R_\text{MTJ}= 1 k \Omega$, DW velocity is $100 \frac{\text{m}}{\text{s}}$ , $J=1.0 \times 10^{11} \frac{\text{A}}{\text{m}^2}$ for SOT switching, $w = 32nm$, $d=4nm$, and $L_\text{mtj}$ is chosen according to Equation \eqref{eq:N_w}. We assume the circuit operates in current mode during VMM operations and during the training/plasticity events, and no additional analog-to-digital conversion (ADC) is needed at the hidden layer due to the all-DW design. However, at the output layer, a Ramp ADC, comparators, and softmax subthreshold circuit are implemented to fully interface with digital labels. Based on our estimates, this peripheral circuitry dominates the overall energy footprint and leads to the following results at 6 bits of ADC accuracy for the three tasks using clustered weights and ana-BP in $\mathcal{W}_s$ : 1.96 $\mu J$ for HAR, 7.41 $\mu J$ for MNIST, and 18.55 $\mu J$ for f-MNIST. Lastly, we parameterize hidden layer dimension and bits ( Fig. \ref{fig:energy}). While energy scales linearly with the system size, it scales quadratically as a function of bits. Since 6 bits of weight precision is workable for Bin-BP and far less suffices for Ana-BP, no blow-up in energy is expected. Future energy efficiencies may be unlocked by further increasing domain wall velocities via material optimization \cite{ajejas2017tuning}, or increasing the efficiency of spin-orbit torque switching for more efficient current-mode inference operations.

\section{Conclusion}
In this work, we have designed and evaluated a learning system which closely draws upon the dynamics of DW-MTJ memory devices to learn efficiently. The major positive result of the work is that current-mode (all DW-MTJ ) internal operation, low bit requirements, and a low number of required updates allow us to achieve learning with $< 20 \mu J$ energy budget at very high speed. The major incomplete aspect of the work is that our accuracy results are still inferior to state-of-the-art deep networks using BP. Our immediate next steps are thus to examine deeper (cascaded) implementations of semi-supervised DW-MTJ systems that may be ML-competitive.


\section*{Acknowledgment}
Sandia National Laboratories is a multimission laboratory managed and operated by NTESS, LLC, a wholly owned subsidiary of Honeywell International Inc., for the U.S. Department of Energy’s National Nuclear Security Administration under contract DE-NA0003525. This paper describes objective technical results and analysis. Any subjective views or opinions that might be expressed in the paper do not necessarily represent the views of the U.S. Department of Energy or the United States Government.



\bibliographystyle{IEEEtran}
\nocite{*} 
\input{bare_conf_PX.bbl} 

%



\end{document}